\documentclass[aps,prl,twocolumn,groupedaddress]{revtex4}

\begin{document}

%\preprint{}

\title{Comment on "Language Trees and Zipping"}


\author{Xiu-Li Wang}
\email[]{wangxiuli@ahu.edu.cn}
%\thanks{}
\affiliation{Departmant of Chinese  Literature and Language Anhui
University  Hefei  Anhui  230039 China }

%\collaboration{}\underline{}

\date{\today}

\begin{abstract}
every encoding has priori information if the encoding represents any
semantic information of the un- verse or object.Encoding means
mapping from the un- verse to the string or strings of digits. The
semantic here is used in the model-theoretic sense or denotation of
the object.if encoding or strings of symbols is the adequate and
true mapping of model or object,and the mapping is recursive or
computable ,the distance between two strings(text)is mapping the
distance between models.We then are able to measure the distance by
computing the distance be- tween the two strings.Oherwise,we may
take a misleading course."language tree"  may not be a family tree
in the sense of historical linguistics.Rather it just means the
similarity
\end{abstract}

\pacs{showpacs}
%\keywords{}

\maketitle

\section{Comment on "Language Trees and Zipping"}
Several statements that Benedetto \sl et al.\rm make in their
Letter~\cite {PhysRevLett.88.048702,PhysRevLett.90.089804}are not
certainly true.First,We claim a statement that Benedetto \sl et
al.\rm. make in their Letter and their reply ~\cite
{PhysRevLett.88.048702,PhysRevLett.90.089804}has mixed strings of
symbols with the objects or models the strings denote.In another
word ,strings of symbols are different from the object or model the
strings denote except when the strings only denote
themselves.Moreover,a statement of the comment on the Letter by
Dmitry V. Khmelev \sl et al.\rm is
inaccurate~\cite{PhysRevLett.90.089803}.That is ,"Notice that the
¡®¡®language tree¡¯¡¯ (LT) diagram [1] does not include the Russian
language (Slavic family of Indo-European family of languages: $288
\times 10^6 $speakers). Our computations show that once Russian is
included, it does not cluster with the other members of the Slavic
group. Obviously, certain Cyrillic alphabet based languages were
left out of the study , which ¡®¡®improves¡¯¡¯ results significantly
and shows that a priori information about the alphabet is being
taken advantage of to achieve the results outlined in their Letter
.".

String of symbols and symbol may self-refer or refer to other
object.When It refer to or denote another object ,we say the object
is model of the string of symbols or meaning (semantics) of the
string of symbols~\cite{Simpsonmodeltheory1998,otto-algorithmic}.The
string of symbols represents the object or the model.Obviously when
It refer to or denote Itself,the meaning or model and the symbol or
string of symbols are the same.The alphabet or text(string of
symbols) are not language.They are symbols or strings of symbols
that just record the language

Clearly ,every encoding has priori information if the encoding
represents any semantic information of the unverse or
object.Encoding means mapping from the unverse to the string or
strings of digits. The semantic here is used in the model-theoretic
sense or denotation of the object .By choosing a string or code that
maps the entities,relation and function in the unverse to symbols
and the relation,function of the symbols ,We encode our knowledge
about the model or object too.If we encode the object by randomly
assigning
the object to a string %that does not map the relation, function in the
%unverse,
everyone or machine can not recognize or get any information about
the unverse or the object without the assignment.For instance,by
isomorphism ,a group is mapped to a group which maintain any
information of the former one such as relations function etc.If the
group is mapped to an other structure randomly ,we can not get any
information about the former one from the latter one without the
mapping,even when we know there exist a mapping from the group to
the structure. We may consider the a logical sentence as the code of
its model.A more concrete example is the binary code of integer.If
the mapping from integer to binary code is random,we can not recover
the integer from its binary code without the mapping.Even the
mapping is not random ,that is, the mapping is recursive or
computable ,we have to make effort to get the information if we know
there exists a mapping that is recursive,or we are unable to get any
information about the integer.Afterall ,the mapping and the model a
string correspond to are priori information that human being
provide.

Therefore,it is true that every encoding  has priori information
which is symbolization(mapping to symbol) of part or all of the
human being's knowledge
about the model.%and the symbolization must be computable.
Even when
 "As for the objection
concerning the coding chosen for our texts, one has to remember that
a zipper ¡®¡®reads¡¯¡¯ the sequences of characters which one inputs
to it, nothing more than this. The idea of comparing languages
written with different alphabets cannot forget this simple
statement. In order to compare languages written with different
alphabets one should, for instance, consider texts written with the
phonetic alphabet. This is the reason for not having included in our
preliminary analysis of the language tree languages such as Chinese,
Greek, Russian, etc.", the phonetic alphabet with which the texts
are written encodes the knowledge of human about the language.

Hence,if the distance that Benedetto \sl et al.\rm define is capable
of the measure of similarity of the compressed text,It at most
measures the similarity between the two text compared .If the
alphabet computationally represent some information of language ,the
distance resulted from the comparison is the measure of the
similarity of
 information of the language.Otherwise It is just the measure of the
 similarity of the text.

 When the compression technique is applied to DNA sequence to cluster
DNA,the distance is just the measure of the similarity.Only under
the presupposition that DNA is mapping of features of creature can
we get some information of creature such as evolution relation or
family tree.

Secondly ,the language tree may not be a family tree .Indo-European
family of languages is not a concept that describe the family
composed of descendants and their
ancestor~\cite{Robinslinguistic1973}.

Many Languages are  descendants of a same archaic one.They are very
similar in spelling,syntax even meaning or semantics when they
inherit or use the same alphabet.Historical linguist compare
language in spelling (phonetics),syntax and meaning to reconstruct
their ancestor.But unfortunately these effort and results are proved
not to be solid or reliable in many cases without data such as
historical text record .Rather,We know that similarity may be
because of type of languages that happen to be similar in some
aspect ,interaction between languages which  is called linguistic
union or being descendant of a same ancient father.There is no
genetic relationship between languages, but they still share
features, and they are spoken in the same region .Balkan linguistic
union or sprachbunds, such as Albanian, Greek, Bulgarian and
Romanian are all IE languages .However, they are not closely
related. Classification of languages may be genetic typological or
areal(linguistic union)~\cite{Robinslinguistic1973}.So,what does the
term ''language tree'' mean?It may not be a family tree in the sense
of historical linguistics.Rather it just means  the
similarity~\cite{Robinslinguistic1973}.By the technique,Benedetto
\sl et al.\rm just show the similarity between the texts ,or the
similarity between the languages that may not be similarity among
members of family only if the similarity between the text (strings
or symbols) is the mapping of   the similarity between the languages
adequately and truly.The language tree is not able to be considered
as a family tree in the sense of historical linguistics.

Thirdly,the distance Benedetto \sl et al.\rm define in their Letter
is similar to the NID definition by Li
Ming~\cite{MingLiVitanyi1997}.As we discuss relation between the
encoding and model above,if encoding or strings of symbols is the
adequate and true mapping of model or object,and the mapping is
recursive or computable ,the distance between two strings(text)is
mapping the distance between models.We then are able to measure the
distance by computing the distance between the two
strings.Oherwise,we may take a misleading course.

There is intention (presupposition) in pure mathematic research that
the mapping from model to string is not considered as a key
question.But application to practical problem may cause trouble or
error.In fact,it has to be solved firstly to decide wether mapping
from model to string or strings contains the information of the
model,although we often do the mapping that is heuristic and valid.
As everyone knows,theory of physics is the "strings",and experiments
of physics is to test or check wether the mapping is valid.The
empirical science may be consider as searching for  and testing
mapping.

\begin{acknowledgments}
Thank Ming-Hui Zhang who works as a faculty in Physics Department of
Anhui University for helpful discussion.
\end{acknowledgments}

\bibliography{reference}

\end{document}